\documentclass[10pt,twocolumn,letterpaper]{article}

\usepackage{cvpr}              

\usepackage[dvipsnames]{xcolor}

\definecolor{cvprblue}{rgb}{0.21,0.49,0.74}
\usepackage[pagebackref,breaklinks,colorlinks,citecolor=cvprblue]{hyperref}

\title{DiffEditor: Boosting Accuracy and Flexibility on Diffusion-based Image Editing}

\author{Chong Mou$^{1}$ \hspace{9pt} Xintao Wang$^{2}$ \hspace{9pt} Jiechong Song$^{1}$ \hspace{9pt} Ying Shan$^{2}$ \hspace{9pt} Jian Zhang$^{1}$\Envelope \\
	\vspace{-0.05cm}
\small$^1$School of Electronic and Computer Engineering, Shenzhen Graduate School, Peking University \hspace{5pt}\\
\small$^2$ARC Lab, Tencent PCG \hspace{5pt}\\
}

\usepackage{amsmath,amsfonts,bm}

\def\eqref#1{equation~\ref{#1}}

\def\1{\bm{1}}

\def\vx{{\bm{x}}}

\DeclareMathAlphabet{\mathsfit}{\encodingdefault}{\sfdefault}{m}{sl}
\SetMathAlphabet{\mathsfit}{bold}{\encodingdefault}{\sfdefault}{bx}{n}

\usepackage{hyperref}
\usepackage{url}
\usepackage{kotex}
\usepackage{graphicx}
\usepackage{amsmath}
\usepackage{amssymb}
\usepackage{amsthm}

\usepackage{multirow}
\usepackage{amsmath}
\usepackage[linesnumbered,ruled,vlined]{algorithm2e}
\usepackage{minitoc}

\usepackage{url}
\usepackage{times}
\usepackage{epsfig}
\usepackage{graphicx}
\usepackage{amsmath}
\usepackage{bbding}
\usepackage{pifont}
\usepackage{wasysym}
\usepackage{amssymb}
\usepackage{bm}
\usepackage{epsfig}
\usepackage{graphicx}
\usepackage{amsmath}
\usepackage{float}
\usepackage{enumitem}
\usepackage{xcolor}
\usepackage{wrapfig}
\usepackage{caption}
\usepackage{makecell}

\begin{document}
\maketitle
\begin{abstract}
Large-scale Text-to-Image (T2I) diffusion models have revolutionized image generation over the last few years. Although owning diverse and high-quality generation capabilities, translating these abilities to fine-grained image editing remains challenging. In this paper, we propose \textbf{DiffEditor} to rectify two weaknesses in existing diffusion-based image editing: (1) in complex scenarios, editing results often lack editing accuracy and exhibit unexpected artifacts; (2) lack of flexibility to harmonize editing operations, e.g., imagine new content. In our solution, we introduce image prompts in fine-grained image editing, cooperating with the text prompt to better describe the editing content. To increase the flexibility while maintaining content consistency, we locally combine stochastic differential equation (SDE) into the ordinary differential equation (ODE)  sampling. In addition, we incorporate regional score-based gradient guidance and a time travel strategy into the diffusion sampling, further improving the editing quality. Extensive experiments demonstrate that our method can efficiently achieve state-of-the-art performance on various fine-grained image editing tasks, including editing within a single image (e.g., object moving, resizing, and content dragging) and across images (e.g., appearance replacing and object pasting). Our source code is released at \url{https://github.com/MC-E/DragonDiffusion}.
\end{abstract}    
\section{Introduction}
\label{sec:intro}
Text-to-image (T2I) diffusion models~\cite{t2i1,ldm,glid,dall-e2} have become the mainstream of image generation, praised for their high-quality and diverse generation capability. The pre-trained T2I models can serve as a good generation prior and can be used in various ways, \textit{e.g.}, image editing. Since the excellent text-to-image ability, numerous diffusion-based image editing methods are implemented based on the text guidance~\cite{p2p, edit1, edit2, selfG,brooks2023instructpix2pix, direct}. However, the generated results of T2I models are usually sensitive to the quality of text~\cite{imageworse}. Therefore, text-guided image editing struggles to achieve fine-grained content manipulation.

\begin{figure}[t]
\centering
\small 
\begin{minipage}[t]{0.9\linewidth}
\centering
\includegraphics[width=1\columnwidth]{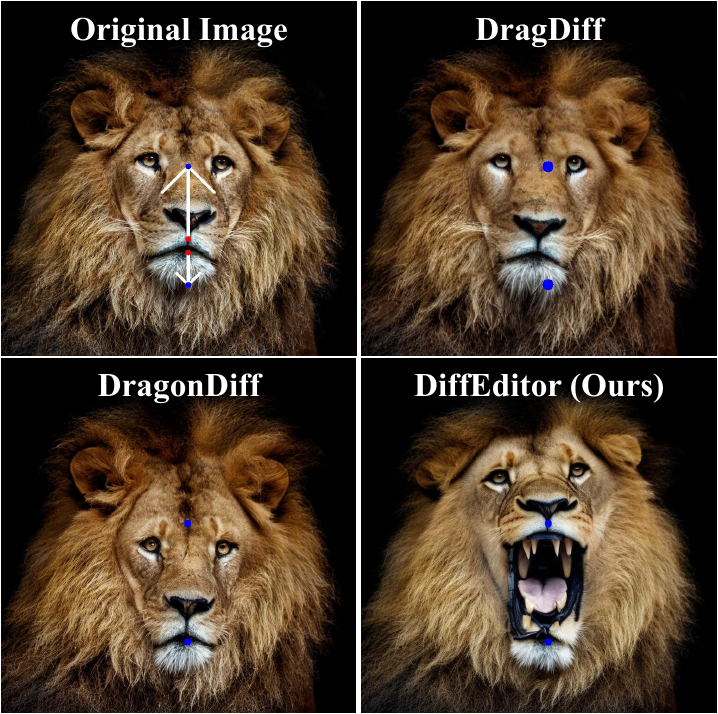}
\end{minipage}
\centering
\caption{Illustration of editing flexibility limitations in DragDiff~\cite{dragdiff} and DragonDiff~\cite{dragondiffusion}, as well as our improvement.
}
\label{motivate} 
\end{figure}

Recently, DragGAN~\cite{draggan} provides a user-friendly way to manipulate the image content by point dragging. However, limited by the capacity of GAN~\cite{gan} models, DragGAN cannot edit general images. Inspired by this interactive editing mode, DragDiff~\cite{dragdiff} and DragonDiff~\cite{dragondiffusion} are proposed based on the pre-trained T2I diffusion model~\cite{ldm}. Empowered by the diverse generation capabilities of the base model, they can perform fine-grained editing on general images. However, their editing process lacks flexibility, as shown in Fig.~\ref{motivate}. Concretely, the image editing operation of transforming a lion from a closed mouth to a widely open mouth conflicts with the LORA~\cite{lora} in DragDiff, resulting in failure. The visual cross-attention designed in DragonDiff also makes it struggle to imagine new content (\textit{e.g.}, mouth) that is not present in the source image, causing failure too. 
In addition, these two methods and most diffusion-based image editing methods employ ordinary differential equations (ODE)~\cite{ddim} solver, a deterministic sampling process. Although ODE can better maintain the consistency between the edited results and the source image, its determinacy also limits flexibility during the editing process. Compared to ODE, stochastic differential equations (SDE)~\cite{diff} is a stochastic sampling process. Some works~\cite{sde1, sde2} study the latent space of SDE for accurate image editing. Unlike these works, we aim to utilize the stochasticity in SDE to improve the flexibility of diffusion-based image editing, as shown in the last image of Fig.~\ref{motivate}.   

Another insight is that although DragDiff and DragonDiff utilize feature correspondence in the pre-trained T2I diffusion model to achieve fine-grained image editing, the role of the text input is ignored in their frameworks. Here, we raise a question: \textit{Does the text have no effectiveness in fine-grained image editing, or is there another more suitable form of text input?} In addition to the text prompt, DALL-E2~\cite{dall-e2} presents a novel attempt to generate images conditioned on the image prompt, \textit{i.e.}, using images to describe images. Subsequently, some multimodal works~\cite{gill,llava,visualinst} and object-customization works~\cite{elite,blipdiffusion,ipadapter} are proposed to support image prompts for more detailed content description. Inspired by these works, we introduce image prompts into the fine-grained image editing process, improving editing quality through more detailed content descriptions. In addition, we combine regional score-based gradient guidance and a time travel strategy into diffusion sampling, which further enhances the editing quality.

In summary, this paper has the following contributions:
\begin{itemize}
    \item We present a novel attempt to introduce the image prompt to fine-grained image editing tasks. In conjunction with the image editing algorithm, this design can provide a more detailed description of the editing content, thus improving the editing quality.
    \item We consider both the flexibility and content consistency of image editing by proposing regional SDE sampling and regional score-based gradient guidance. We also introduce a time travel strategy in diffusion-based image editing to improve the editing quality further.
    \item Extensive experiments demonstrate that our method can achieve state-of-the-art performance on various fine-grained image editing tasks (\textit{i.e.}, content dragging, object moving, resizing, pasting, and appearance replacing, as shown in Fig.~\ref{fig:teaser}) with attractive complexity.
\end{itemize}

\section{Related Work}
\label{sec:relat}
\subsection{Diffusion Model}
\begin{figure*}[t]
\centering
\begin{minipage}[t]{\linewidth}
\centering
\includegraphics[width=1\columnwidth]{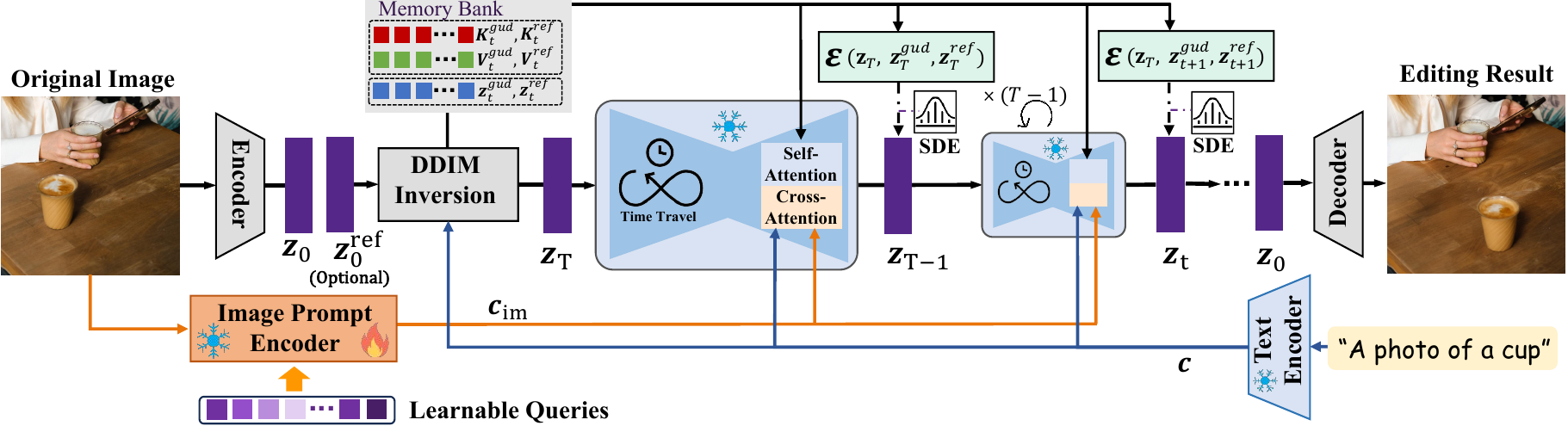}
\end{minipage}
\centering
\caption{Overview of our proposed DiffEditor, which is composed of a trainable image prompt encoder and a diffusion sampling with editing guidance that does not require training.
}
\label{network} 
\end{figure*}
Diffusion model~\cite{diff} is a thermodynamics-driven~\cite{phy1,phy2} algorithm, including a diffusion process and a reverse process. In the diffusion process, an image $\mathbf{x}_0$ is gradually added Gaussian noise as $q(\mathbf{x}_{t}| \mathbf{x}_0)=\mathcal{N}(\sqrt{\alpha_t}\mathbf{x}_0, (1-\alpha_t)\mathbf{I})$, where $\alpha_t$ linearly decreases from 1 to a sufficiently small number to encourage $\mathbf{x}_T\sim \mathcal{N}(0,\mathbf{I})$. The reverse process is to iteratively recover $\mathbf{x}_0$ from $\mathbf{x}_T$ by training a denoiser, conditioned on the current noisy image $\mathbf{x}_{t}$ and time step $t$:
\begin{equation}
\label{loss}
    \mathbb{E}_{\mathbf{x}_{0},t, \boldsymbol{\epsilon}_t \sim \mathcal{N}(0,1)}\left[ ||\boldsymbol{\epsilon}_t-\epsilon_{\boldsymbol{\theta}}(\mathbf{x}_{t},t)||_2^2\right],
\end{equation}
where $\epsilon_{\boldsymbol{\theta}}$ is the function of the denoiser. DDIM~\cite{ddim} defines the diffusion sampling as $q(\mathbf{x}_{t-1}|\mathbf{x}_t, \mathbf{x}_0)=\mathcal{N}(\sqrt{\alpha_{t-1}}\mathbf{x}_0+\sqrt{1-\alpha_{t-1}-\delta_t^2}\cdot \frac{\mathbf{x}_t-\sqrt{\alpha_t\mathbf{x}_0}}{\sqrt{1-\alpha_t}}, \alpha_t^2\mathbf{I})$, which is a non-Markovian process and can be formulated as:
\begin{flalign}
\label{ddim_reverse}
\footnotesize
\begin{split}
    &\vx_{t-1}=\\
    &\sqrt{\alpha_{t-1}} \underbrace{\frac{\vx_{t}-\sqrt{1-\alpha_{t}} \bm{\epsilon}_{\theta}^{t}\left(\vx_{t}\right)}{\sqrt{\alpha_{t}}}}_{\text {"predicted } \vx_{0} \text { " }}+ \underbrace{\sqrt{1-\alpha_{t-1}-\sigma_{t}^{2}} \cdot \bm{\epsilon}_{\theta}^{t}\left(\vx_{t}\right)}_{\text {"direction pointing to } \vx_{t} \text { " }}+ \underbrace{\sigma_{t} \bm{\epsilon}}_{\text {"noise" }},
\end{split}  \hspace{4cm}
\end{flalign}
where $\sigma_t = \eta \sqrt{\left(1-\alpha_{t-1}\right) /\left(1-\alpha_{t}\right)} \sqrt{1-\alpha_{t} / \alpha_{t-1}}$. When $\eta=1$ for all $t$, it becomes DDPM~\cite{diff}, \textit{i.e.}, a stochastic differential equation (SDE). As $\eta=0$, the sampling process becomes deterministic, \textit{i.e.}, an ordinary differential equation (ODE). Most diffusion-based image editing works rely on ODE to achieve better content consistency. \cite{sde1, sde2} explore SDE in diffusion-based image editing.

Most current works focus on conditional diffusion generation, such as text conditions~\cite{glid,ldm}, which have greatly revolutionized the community of image generation. Although promising T2I generation quality is achieved, the generated results are sensitive to text quality and usually require tedious prompt design~\cite{imageworse}. In addition to text condition, DALL-E2~\cite{dall-e2} presents the first attempt to generate images guided by image prompts. ELITE~\cite{elite}, Bilp-Diffusion~\cite{blipdiffusion}, and IP-Adapter~\cite{ipadapter} present the learning of image prompts for object customization. However, the effectiveness of image prompts in fine-grained image editing has hardly been studied.

\subsection{Image Editing}
The primary objective of image editing is to manipulate the content of a given image in a controlled manner. Previous methods~\cite{abdal2019image2stylegan, abdal2020image2stylegan++, alaluf2022hyperstyle} usually invert images into the latent space of GANs~\cite{gan} and then edit the image by manipulating latent vectors. Recently, DragGAN~\cite{draggan} presents a point-dragging formulation for fine-grained image editing. However, limited by the capability of GANs, these methods have weaknesses in model generalization and image quality. Motivated by the success of text-to-image diffusion models~\cite{ldm}, various text-guided image editing methods~\cite{avrahami2022blended, p2p,kawar2023imagic,meng2021sdedit,brooks2023instructpix2pix} are proposed. The commonly used editing strategies are (1) adding noise and then denoising with target description~\cite{kawar2023imagic, valevski2023unitune, kwon2022diffusion, meng2021sdedit, masactrl}; (2) using cross-attention maps as an editing medium~\cite{p2p, selfG, direct}; (3) using text as editing instructions~\cite{brooks2023instructpix2pix}. However, the correspondence between the text and image in T2I models is weak, making it difficult to achieve fine-grained image editing. Recently, DragDiff~\cite{dragdiff} and DragonDiff~\cite{dragondiffusion} achieve fine-grained image editing based on the feature correspondence~\cite{dift} in the pre-trained StableDiffusion (SD)~\cite{ldm}. Specifically, DragDiff uses LORA~\cite{lora} to maintain content consistency and optimizes the latent $\mathbf{z}_t$ in a specific diffusion step. DragonDiff is built based on the score-based~\cite{score} gradient guidance~\cite{clf} and a visual cross-attention design for drag-style image editing without model tuning. 
\section{Method}
\subsection{Preliminary: Score-based Editing Guidance}
From the continuous perspective of score-based diffusion~\cite{score, score1}, the external condition $\mathbf{y}$ can be combined in a conditional score function, \textit{i.e.}, $\nabla_{\mathbf{x}_t} \log q(\mathbf{x}_t | \mathbf{y})$, to sample from a more enriched distribution. The conditional score function can be further decomposed as:
\begin{flalign}
\label{fn_score}
\begin{split}
    \nabla_{\mathbf{x}_t} \log q(\mathbf{x}_t | \mathbf{y}) &=  \nabla_{\mathbf{x}_t} \log\left(\frac{q(\mathbf{y}|\mathbf{x}_t)q(\mathbf{x}_t)}{q(\mathbf{y})}\right)\\ &\propto \nabla_{\mathbf{x}_t} \log q(\mathbf{x}_t)+\nabla_{\mathbf{x}_t} \log q(\mathbf{y}|\mathbf{x}_t),
\end{split}
\end{flalign}
where the first term is the unconditional denoiser, \textit{i.e.}, $\bm{\epsilon}_{\theta}^{t}(\mathbf{x}_t)$. The second term refers to the conditional gradient produced by an energy function $\mathcal{E}(\mathbf{x}_t, \mathbf{y}) = \log q(\mathbf{y}|\mathbf{x}_t)$, measuring the distance between current state $\mathbf{x}_t$ and condition $\mathbf{y}$. Here, we reformulate Eq.~\ref{fn_score} as:
\begin{equation}
\label{fn_score_my}
    \tilde{\bm{\epsilon}}_{\theta}^t(\mathbf{x}_t)=\bm{\epsilon}_{\theta}^t(\mathbf{x}_t) + \eta\cdot\nabla_{\mathbf{x}_t} \mathcal{E}(\mathbf{x}_t, \mathbf{y}),
\end{equation}
where $\eta$ refers to the learning rate. Recently, Self-Guidance~\cite{selfG} and DragonDiff~\cite{dragondiffusion} convert image editing operations into gradient guidance for image editing tasks. The energy function $\mathcal{E}$ in Self-Guidance is built based on the correspondence~\cite{p2p} between image and text features. DragonDiff constructs the energy function based on image feature correspondence~\cite{dift} in pre-trained SD, which can achieve more accurate drag-style editing tasks. In this paper, we aim to boost the accuracy and flexibility of diffusion-based image editing with the DragonDiff framework.

\subsection{Overview}
An overview of our image editing pipeline is presented in Fig.~\ref{network}. Specifically, given an image $\mathbf{x}_0$ to be edited, we first employ it as image prompts and use an image prompt encoder to embed it. These image embeddings cooperate with text embeddings to form a better description to guide the diffusion process. Then we use DDIM inversion~\cite{ddim} to transform $\mathbf{x}_0$ into a Gaussian distribution $\mathbf{z}_T$ in the latent space of the pre-trained SD~\cite{ldm}. If the reference image $\mathbf{x}_0^{ref}$ exists (\textit{i.e.}, in appearance replacing and object pasting), it will also be involved in the inversion. In this process, we follow the design in DragonDiff~\cite{dragondiffusion} to store some intermediate features ( $\mathbf{K}_{t}^{gud}, \mathbf{V}_{t}^{gud}, \mathbf{K}_{t}^{ref}, \mathbf{V}_{t}^{ref}$) and latent ($\mathbf{z}_t^{gud},\mathbf{z}_t^{ref}$) at each time step in a memory bank, which is used to guide subsequent image editing. Note that "gud" and "ref" represent the information of the source and reference image in the inversion process, respectively. In the subsequent generation sampling, we step forward with the cooperation of score-based editing guidance, visual cross-attention, and image prompt. In this process, some elaborately designed strategies (\textit{e.g.}, regional gradient guidance, regional SDE, and time travel) enhance the editing further.

\begin{figure}[t]
\centering
\begin{minipage}[t]{\linewidth}
\centering
\includegraphics[width=1\columnwidth]{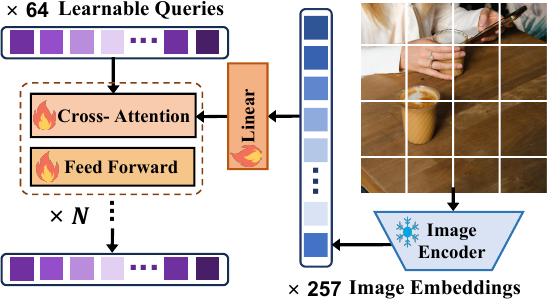}
\end{minipage}
\centering
\caption{
Illustration of the design of our image prompt encoder.
}
\label{ibm} 
\end{figure}

\subsection{Content Description with Image Prompt}
Although several fine-grained image editing methods~\cite{dragdiff, dragondiffusion} are based on the T2I diffusion model, the role of prompts is ignored as a simple description. Compared to text prompts, image prompts~\cite{dall-e2, blipdiffusion, ipadapter} can provide a more detailed content description. In this paper, we find that the image prompt can improve the quality of fine-grained image editing, especially in some complex scenarios. 

Inspired by IP-Adapter~\cite{ipadapter}, the architecture of our image prompt encoder is shown in Fig.~\ref{ibm}. Concretely, given an input image $\mathbf{x}_0$, the pre-trained CLIP~\cite{clip} image encoder embed it to 257 tokens. Then, a linear layer is used to adjust the channel dimension, and a QFormer~\cite{li2023blip} (without self-attention layer) module is employed to adjust the token numbers to 64 by 64 learnable queries. The QFormer module consists of N (8 by default) submodules, each composed of a cross-attention layer and a feed-forward network (FN). 64 learnable queries serve as queries to extract information from 257 image tokens that act as keys and values. Finally, 257 image tokens are composed into 64 embedding tokens ($\mathbf{c}_{im}$), and then they are input into the same cross-attention module as text tokens ($\mathbf{c}$) in the SD. To build classifier-free guidance~\cite{ho2022classifier} like the text condition, the conditional and unconditional image prompts are jointly trained by randomly dropping (\textit{i.e.}, set image to zero) during training. Finally, image tokens and text tokens are processed separately with the query $\mathbf{Q}$ in the cross-attention module, and the results are added together:
\begin{equation}
\footnotesize
    \text{Att}(\mathbf{Q}, \mathbf{K}^{'}, \mathbf{V}^{'}, \mathbf{K}^{''}, \mathbf{V}^{''})=\mathcal{S}(\frac{\mathbf{Q}(\mathbf{K}^{'})^T}{\sqrt{d}})\mathbf{V}^{'}+\gamma\cdot\mathcal{S}(\frac{\mathbf{Q}(\mathbf{K}^{''})^T}{\sqrt{d}})\mathbf{V}^{''},
\end{equation}
where $(\mathbf{K}^{'}, \mathbf{V}^{'})$ and $(\mathbf{K}^{''}, \mathbf{V}^{''})$ refer to the keys and values from the text and image prompt, respectively. $\gamma$ is a weight to balance these two terms. $\mathcal{S}$ is the function of Softmax. Note that in tasks with reference images (\textit{i.e.}, object pasting and appearance replacing), $\mathbf{K}^{''}$ and $\mathbf{V}^{''}$ are formed by the concatenation of image tokens from the source image and the reference image. During training, we fix the parameters in the pre-trained SD and CLIP image encoder, and we only optimize the linear embedding and QFormer by $\mathcal{L}_2$ loss:  
\begin{equation}
\label{loss_qf}
    \mathbb{E}_{\mathbf{x}_{0},t, \boldsymbol{\epsilon}_t \sim \mathcal{N}(0,1)}\left[ ||\boldsymbol{\epsilon}_t-\epsilon_{\boldsymbol{\theta}}^t(\mathbf{z}_{t},\mathbf{c},\mathbf{c}_{im})||_2^2\right].
\end{equation}
After a single training, this module can be integrated into pre-trained SD for various image editing tasks, as demonstrated in this paper.

\begin{figure}[t]
\centering
\begin{minipage}[t]{\linewidth}
\centering
\includegraphics[width=1\columnwidth]{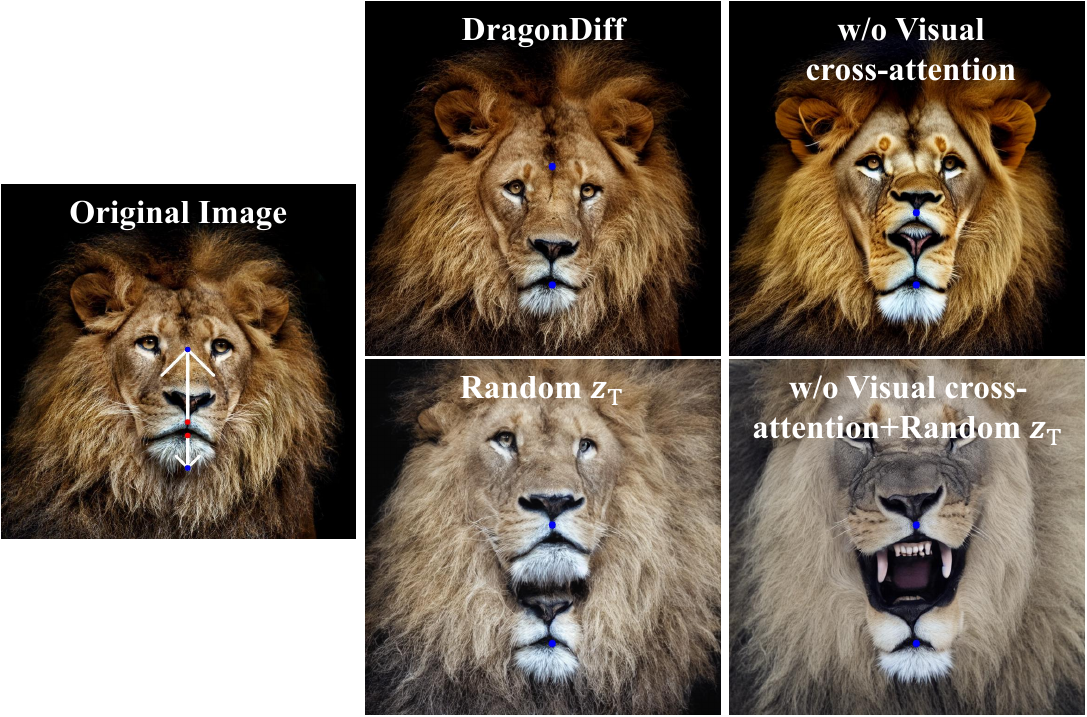}
\end{minipage}
\centering
\caption{The impact of different components on the editing flexibility of DragonDiff~\cite{dragondiffusion}.}
\label{sde} 
\end{figure}

\subsection{Sampling with Regional SDE}
Maintaining consistency between editing results and original images is a great challenge in fine-grained image editing. Most methods adopt a deterministic sampling process (ODE) and utilize DDIM inversion for sampling initialization. In addition, DragDiff~\cite{dragdiff} uses LORA~\cite{lora} to constrain the output content, and DragonDiff~\cite{dragondiffusion} uses visual cross-attention to maintain content consistency. However, these strategies also compromise editing flexibility, \textit{e.g.}, hindering the imagination of new content to harmonize editing operation as shown in Fig.~\ref{motivate}. Our further experiments on DragonDiff show that reducing the content consistency strength can improve editing flexibility. As seen in Fig.~\ref{sde}, the editing flexibility is improved when we randomly initialize the sampling starting point $\mathbf{z}_T$ or remove visual cross-attention. 
When we apply both reductions, the editing objective can be achieved flexibly, but the content consistency is severely compromised. Therefore, in this paper, we explore how to improve editing flexibility without significantly impacting content consistency. In the sampling process (\textit{i.e.}, Eq.~\ref{ddim_reverse}) of DragonDiff, $\sigma_{t}=0$, which is a deterministic ODE sampling. This leads to the final result being highly dependent on $\mathbf{z}_T$ and the information injected by visual cross-attention. Our solution is to introduce randomness (\textit{i.e.}, $\sigma_{t}>0$) during the sampling process, while this randomness is controlled within local editing areas and specific time intervals. Here, we use $\mathbf{z}_{t-1}=\mathcal{F}(\mathbf{z}_t;\sigma_t)$ to simplify Eq.~\ref{ddim_reverse}. Our regional SDE sampling is defined as:
\begin{equation}
\label{fn_sde}
\begin{split}
&\mathbf{z}_{t-1}=\mathbf{m}_{edit}\cdot \mathcal{F}(\mathbf{z}_t;\eta_1(t))+(1-\mathbf{m}_{edit})\cdot \mathcal{F}(\mathbf{z}_t;\eta_2(t)),\\
     &(\eta_{1}(t), \eta_{2}(t)) = \left\{ \begin{array}{ll}
    (0.4, 0.2),                    & t\in \tau_{SDE} \\
    (0, 0),     & t\notin \tau_{SDE}\\
\end{array}
\right.
\end{split}
\end{equation}
where $\mathbf{m}_{edit}$ locates the editing area. $\tau_{SDE}$ is the time interval for applying regional SDE. After using this sampling strategy, we can accurately inject flexibility to produce satisfactory results, as shown in the last image of Fig.~\ref{motivate}.

\begin{figure}[t]
\centering
\small 
\begin{minipage}[t]{\linewidth}
\centering
\includegraphics[width=1\columnwidth]{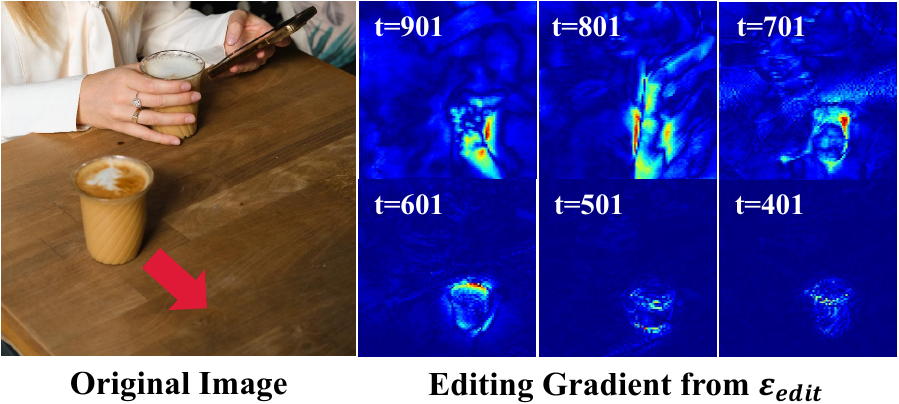}
\end{minipage}
\centering
\caption{
Visualization of the editing gradient from $\mathcal{E}_{edit}$ in different sampling steps.
}
\label{gradient}
\end{figure}

\subsection{Editing with Gradient Guidance}
\label{sec_gradient}
\noindent \textbf{Regional gradient guidance}. In DragonDiff~\cite{dragondiffusion}, the energy function $\mathcal{E}$ consists of two parts, \textit{i.e.}, editing $\mathcal{E}_{edit}$ and content consistency $\mathcal{E}_{content}$. 
Although their target areas are independent of each other, the scope of the gradient guidance they generate is global and overlapping, resulting in mutual interference.
Concretely, in Fig.~\ref{gradient}, we visualize the editing gradient produced by $\mathcal{E}_{edit}$ in the object moving task. As can be seen, the gradient guidance gradually converges to the editing area as the diffusion sampling proceeds. During this process, there are some activations outside the editing area, and these imprecise activations can affect the content consistency in these unedited areas (details are presented in Sec.~\ref{sec_abs}). To rectify this weakness, we use the editing region mask $\mathbf{m}_{edit}$ to locally combine $\mathcal{E}_{edit}$ and $\mathcal{E}_{content}$. Finally, the conditional term in Eq.~\ref{fn_score} is defined as:
\begin{equation}
\small
\label{l_gradient}
    \nabla_{\mathbf{z}_t} \log q(\mathbf{y}|\mathbf{z}_t) = \mathbf{m}_{edit}\cdot \nabla_{\mathbf{x}_t}\mathcal{E}_{edit} + (1-\mathbf{m}_{edit})\cdot \nabla_{\mathbf{x}_t}\mathcal{E}_{content},
\end{equation}
where $\mathbf{y}$ is the editing target. During the sampling, we only add guidance in the first $n$ time steps. 

\begin{algorithm}[t]
\caption{Proposed DiffEditor}
\label{aug}
\textbf{Require}:

\hspace{0.01cm}pre-trained SD~\cite{ldm} $\epsilon_{\boldsymbol{\theta}}$; image to be edited $\mathbf{x}_{0}$; mask of the editing region $\mathbf{m}_{edit}$; gradient-guidance steps $n$; time interval $\tau_{SDE}$ for SDE; time travel interval $\tau_{TT}$ and the internal iterations $U$.

\textbf{Initialization}:

    \hspace{0.01cm} (1) Compute text embedding $\mathbf{c}$ and image image embedding $\mathbf{c}_{im}$
    
    \hspace{0.01cm} (2) invert $\mathbf{x}_0$ to $\mathbf{z}_{T}^{gen}$ and build the memory bank

 \For{$t=T,\ \ldots,\ 1$}{
\eIf{$t\in \tau_{TT}$ and $t\%2==0$}{$U_{cur}=U$}{$U_{cur}=1$}
 
 \For{$u=1,\ \ldots,\ U_{cur}$}{
noise prediction: $\hat{\boldsymbol{\epsilon}}_t=\epsilon_{\boldsymbol{\theta}}(\mathbf{z}_{t}, t,\mathbf{c},\mathbf{c}_{im})$;

\eIf{$T-t<n$ and $t\%2==0$}{compute $\mathcal{E}_{edit}$ and $\mathcal{E}_{content}$ by \cite{dragondiffusion};

compute $\nabla_{\mathbf{z}_t} \log q(\mathbf{y}|\mathbf{z}_t)$ by Eq.~\ref{l_gradient};

inject gradient guidance by Eq.~\ref{fn_score};

compute $\mathbf{z}_{t-1}$ by Eq.~\ref{fn_sde};}{compute $\mathbf{z}_{t-1}$ by Eq.~\ref{ddim_reverse} ($\sigma_t=0$);
}
\uIf{$U_{cur}>1$}{$\mathbf{z}_t=\frac{\mathbf{z}_{t-1}-\sqrt{1-\bar{\alpha}_{t}}\epsilon_{\boldsymbol{\theta}}(\mathbf{z}_t,t,\mathbf{c})}{\sqrt{\bar{\alpha}_{t}}}$}
}
 }

\hspace{0.01cm} $\mathbf{x}_0 = Decoder(\mathbf{z}_0)$;

\textbf{Output:} $\mathbf{x}_0$
\end{algorithm}

\begin{table*}[t]
\caption{Quantitative evaluation on face manipulation with 68 and 17 points. The accuracy is calculated by MSE distance between edited points and target points. The initial distance (\textit{i.e.}, \textit{57.19} and \textit{36.36}) is the upper bound, without editing.
FID~\cite{fid} is utilized to quantize the editing quality of different methods. The time complexity is computed on the ‘1 point’ dragging.}
\small
\centering
\begin{tabular}{c | c c c c c c}
\hline
 & \makecell[c]{Preparing\\complexity$\downarrow$} & \makecell[c]{Inference\\complexity$\downarrow$} & \makecell[c]{Unaligned\\face} & \makecell[c]{17 Points$\downarrow$\\From 57.19} & \makecell[c]{68 Points$\downarrow$\\ From 36.36} & \makecell[c]{FID$\downarrow$\\17/68 points} \\
\hline
UserControllableLT~\cite{user} & \textbf{1.1}s & \textbf{0.04}s & \XSolidBrush & 32.32 & 24.15 & 51.20/50.32 \\
DragGAN~\cite{draggan} & 50.22s & \underline{6.28s} & \XSolidBrush & \textbf{15.96} & \textbf{10.60} & 39.27/39.50\\
DragDiff~\cite{dragdiff} & 42.37s & 19.52s & \Checkmark & 22.95 & 17.32 & 38.06/36.55\\
DragonDiff~\cite{dragondiffusion} & \underline{3.53s} & 15.00s & \Checkmark & 18.51 & 13.94 & \underline{35.75}/\underline{34.58}\\
DiffEditor (Ours) & \underline{3.53s} & 13.88s & \Checkmark &  \underline{17.05}& \underline{11.52} & \textbf{33.10}/\textbf{33.02}\\
\hline
\end{tabular}
\label{tb_cp_face}
\end{table*}

\begin{figure*}[t]
\centering
\small 
\begin{minipage}[t]{0.9\linewidth}
\centering
\includegraphics[width=1\columnwidth]{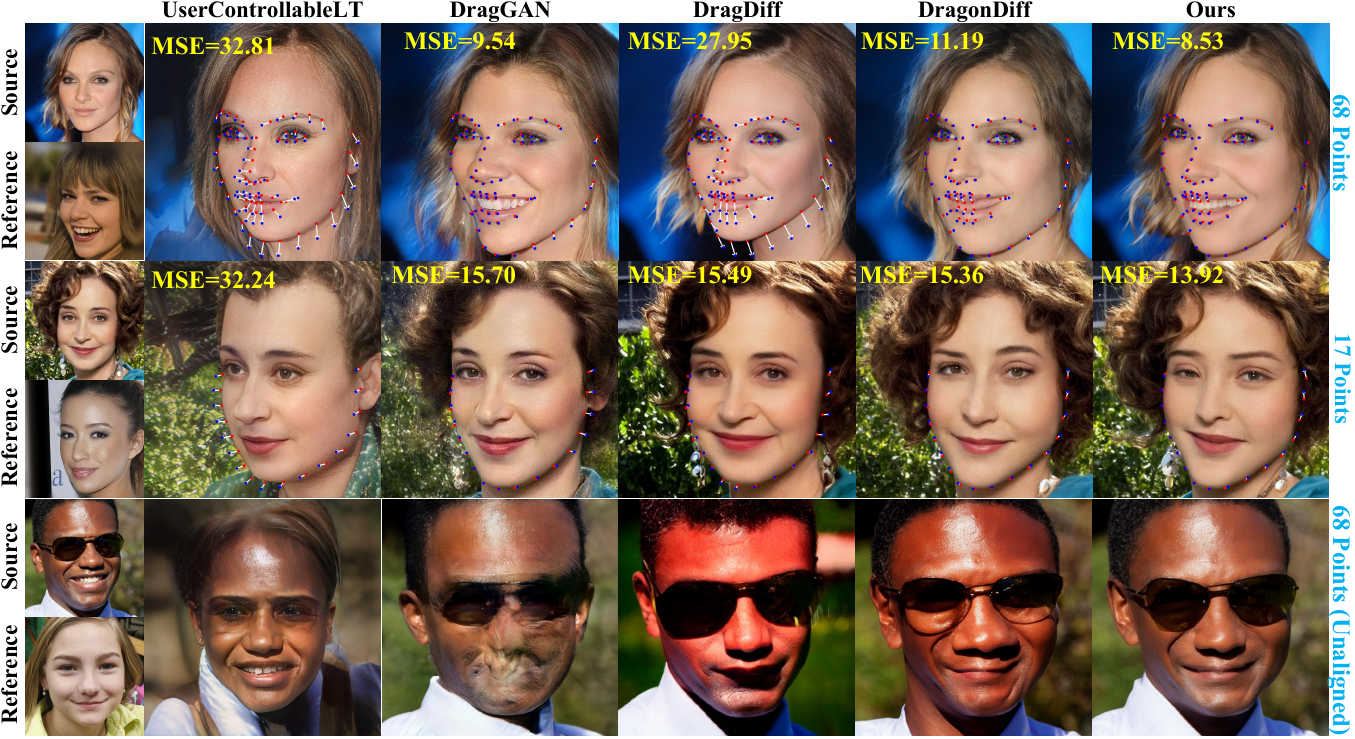}   
\end{minipage}
\centering
\caption{Qualitative comparison between our DiffEditor and other methods in face manipulation. The current and target points are labeled with \textcolor{red}{red} and \textcolor{blue}{blue}. The white line indicates distance. The MSE distance between the result and the target is labeled in yellow.}
\label{fig_cmp_face} 
\end{figure*}

\noindent \textbf{Time travel}. DragDiff~\cite{dragdiff} treats $\mathbf{z}_t$ as a learnable parameter and iteratively optimizes it within a diffusion step $t$. In contrast, DragonDiff~\cite{dragondiffusion} incorporates score-based gradient guidance into each sampling step, \textit{i.e.}, Eq.~\ref{fn_score_my}. However, applying editing guidance through Eq.~\ref{fn_score_my} once at each sampling step lacks refinement for editing, especially in some complex scenarios. \textit{Can we combine the advantages of DragDiff and DragonDiff to build recurrent guidance in the score-based diffusion~\cite{score}?} To address this issue, we build time travel to perform rollbacks, \textit{i.e.}, $\mathbf{z}_t\leftarrow \mathbf{z}_{t-1}$, during the sampling process. This strategy has been empirically shown to inhibit the generation of disharmonious results when solving hard generation tasks~\cite{freedom, lugmayr2022repaint}. However, the rollback strategy (\textit{i.e.}, $\mathbf{z}_t\sim \mathcal{N}(\sqrt{1-\beta_{t-1}}\mathbf{z}_{t-1}, \beta_{t-1}\mathbf{I})$) in these works is not suitable in fine-grained image editing tasks. This is because random noise $\mathbf{I}$ can introduce significant uncertainty, undermining the content consistency of editing results. To ensure the accuracy of rollback, we use deterministic DDIM inversion~\cite{ddim} to roll back $\mathbf{z}_{t-1}$ to $\mathbf{z}_{t}$. 
During sampling, the time travel is performed $U$ (3 in our design) times for each guidance step in a time interval $\tau_{TT}$. 

Due to the guidance enhancement from regional guidance and time travel, we can achieve editing with fewer guidance time steps, \textit{i.e.}, we introduce gradient guidance every two time steps in sampling. Finally, the algorithm logic of our DiffEditor is defined in Alg.~\ref{aug}.

\begin{figure*}[t]
\centering
\small 
\begin{minipage}[t]{\linewidth}
\centering
\includegraphics[width=1\columnwidth]{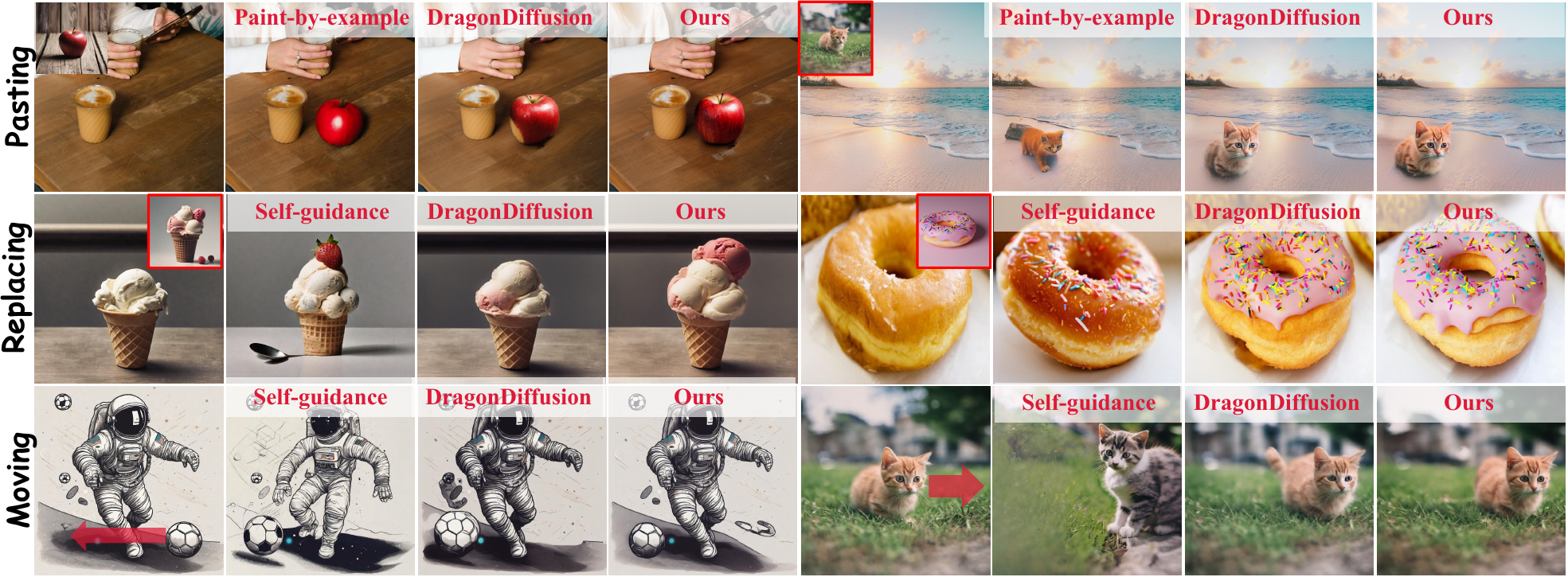}
\end{minipage}
\centering
\caption{
Visual comparison between our method and other methods on appearance replacing, object pasting and object moving tasks. 
}
\label{cmp_dragon} 
\end{figure*}

\section{Experiments}
\subsection{Implementation Details}
We choose Stable Diffusion V1.5~\cite{ldm} as the base model for image editing. During image prompt training, we use the training data from LAION~\cite{laion} and process the image resolution to $512\times 512$. We choose Adam~\cite{adam} optimizer with an initial learning rate of $1\times 10^{-5}$. The batch size during training is set as 16. The training process iterates $1\times 10^6$ steps on 4 NVIDIA A100 GPUs. We use the same embedding module to process image prompts in different applications. The inference adopts DDIM sampling with 50 steps, and we set the classifier-free guidance scale as 5. 

\subsection{Comparison}

\noindent \textbf{Time complexity.} We divide the time complexity of different methods into preparing and inference stages. The preparing stage involves Diffusion/GAN inversion and model tuning. The inference stage generates the editing result from latent representation. The time complexity for each method is tested on one point dragging, with the image resolution being $512\times 512$. All times are tested on an NVIDIA A100 GPU with Float32 precision. The results in Tab.~\ref{tb_cp_face} present the attractive preparing complexity of our method, and the inference complexity is lower than existing diffusion-based methods, \textit{i.e.}, DragDiff and DragonDiff. 

\begin{table}[t]
\caption{Quantitative evaluation on object pasting, object moving, and appearance replacing. The result is calculated by CLIP~\cite{clip} $\uparrow$ distance between editing results and target descriptions.}
\small
\centering
\setlength{\tabcolsep}{4.2pt}
\begin{tabular}{c | c c c}
\toprule
& Pasting & Moving & Replacing\\
\hline
Pain-by-example & 0.265 & - & - \\
Self-Guidance & - & 0.246 & 0.243\\
DragonDiff & 0.260 & 0.282 & 0.263\\
Ours & 0.274 & 0.288 & 0.281\\
\toprule
\end{tabular}
\label{tb_cp_more}
\end{table}

\begin{figure}[t]
\centering
\small 
\begin{minipage}[t]{\linewidth}
\centering
\includegraphics[width=1\columnwidth]{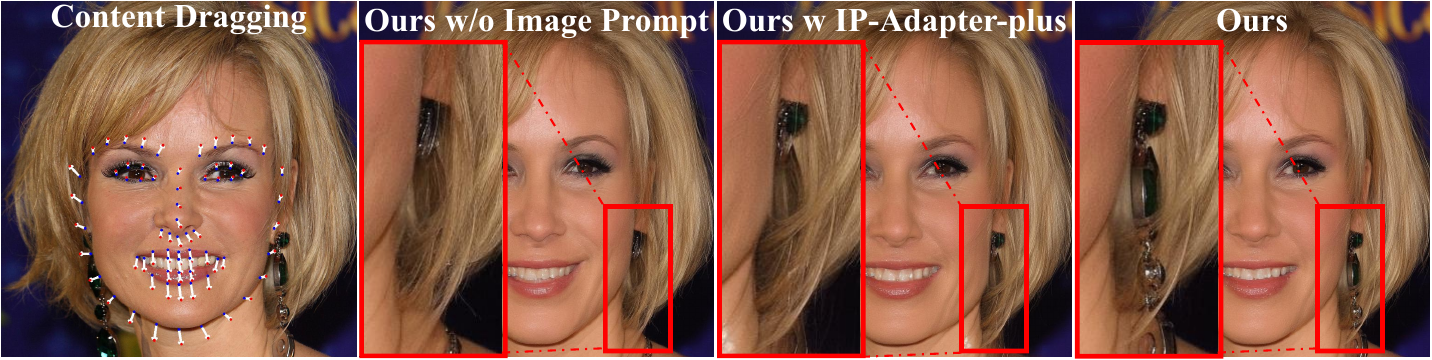}
\end{minipage}
\centering
\caption{
Visual comparison between our image prompt and IP-Adapter-plus.
}
\label{cp_ip}
\end{figure}

\begin{figure}[t]
\centering
\small 
\begin{minipage}[t]{\linewidth}
\centering
\includegraphics[width=1\columnwidth]{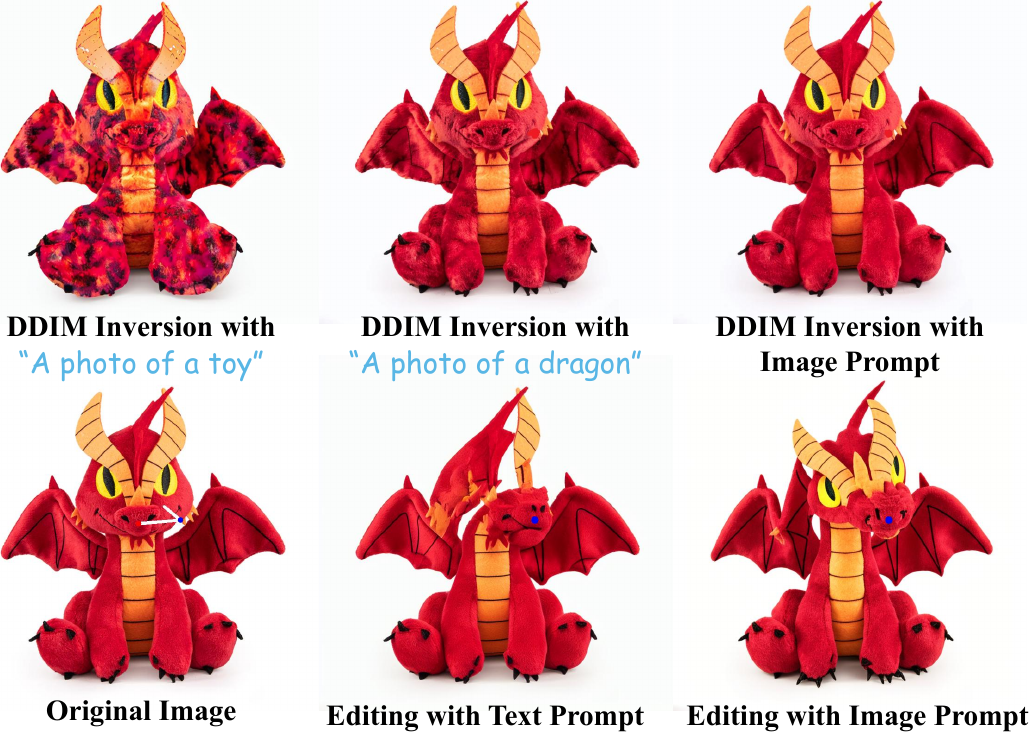}
\end{minipage}
\centering
\caption{
The first and second rows show the effectiveness of the image prompt in DDIM inversion and image editing, respectively.
}
\label{abs_ibm}
\end{figure}

\noindent \textbf{Performance.} First, we evaluate our method on content dragging by comparing it with some well-known GAN-based methods (\textit{i.e.}, UserControllableLT~\cite{user}, DragGAN~\cite{draggan}) and recent diffusion-based methods (\textit{i.e.}, DragDiff~\cite{dragdiff}, DragonDiff) on the keypoint-based face manipulation. We used the same test set as DragonDiff, \textit{i.e.}, 800 aligned faces from the CelebA-HQ~\cite{celeba} training set. We evaluate the editing performance under 17-point guidance and 68-point guidance. To quantify editing accuracy, we calculated the MSE distance between the landmarks of the edited result and the target landmarks. In addition, we calculate FID~\cite{fid} between the editing results and the CelebA-HQ training set to represent the image quality. The quantitative comparison is presented in Tab.~\ref{tb_cp_face}. One can see that our method has a significant improvement in accuracy and generation quality compared to other diffusion-based methods, achieving comparable editing accuracy to DragGAN. Although DragGAN has higher editing accuracy on aligned faces, its base model is specifically trained for aligned faces and cannot edit general faces, as shown in the last row of Fig.~\ref{fig_cmp_face}. The qualitative comparison in Fig.~\ref{fig_cmp_face} shows that our method has high editing accuracy and content consistency while maintaining good flexibility. For example, in the case where teeth need to be imagined, our DiffEditor can produce more natural results. In contrast, DragDiff and DragonDiff have difficulties in imagining new content.

In addition to content dragging, we also compare with Paint-by-example~\cite{paint-by} in object pasting, and we compare our method with Self-Guidance~\cite{selfG} and DragonDiff in object moving and appearance replacing tasks. The results are presented in Fig.~\ref{cmp_dragon}. As can be seen, although the specially trained Paint-by-example can naturally integrate objects into an image, it is difficult to maintain the original object identity. Our method performs better in object identity and has richer texture details than DragonDiff. In object moving and appearance replacing tasks, text-guided Self-Guidance lacks consistent constraints, making editing results deviate from the original image. The editing accuracy of DragonDiff still has room for improvement, \textit{e.g.}, color and details. In comparison, our method has better content consistency and editing accuracy. For quantization, we compute the CLIP~\cite{clip} distance between the edited results and the target description. We select 16 editing samples for each task. The results in Tab.~\ref{tb_cp_more} demonstrate the promising performance of our method.

\noindent \textbf{Discussion between our image prompt and IP-Adapter.} As mentioned above, there are several methods proposed to use images as prompts to provide more accurate and customized descriptions for the generated results, such as IP-Adapter~\cite{ipadapter}. However, most of these methods focus on object customization, and overemphasis on detail description will compromise their performance. Therefore, IP-Adapter compresses the image into a small number of tokens to avoid detail descriptions. This paper studies introducing image prompts into fine-grained image editing. We use the Q-Former structure to map the image into 64 tokens to enhance the detail expression ability. Fig.~\ref{cp_ip} shows that IP-Adapter not ideal for direct insertion into fine-grained image editing tasks due to the lack of detail description. Our method can enhance the consistency of texture details between the edited result and the original image.  


\begin{figure}[t]
\centering
\small 
\begin{minipage}[t]{\linewidth}
\centering
\includegraphics[width=1\columnwidth]{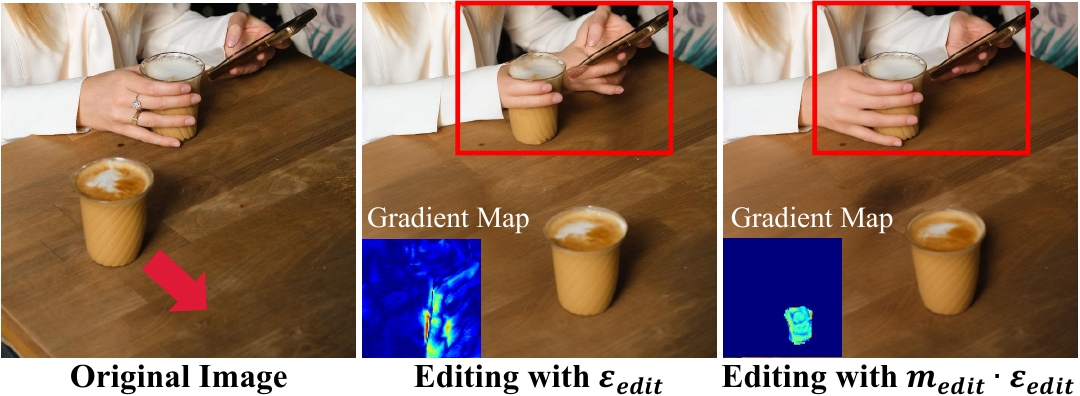}
\end{minipage}
\centering
\caption{
Effectiveness of the regional gradient guidance. In experiment, we remove the content consistency gradient $\mathcal{E}_{content}$.
}
\label{abs_gradient} 
\end{figure}

\begin{figure}[t]
\centering
\small 
\begin{minipage}[t]{0.9\linewidth}
\centering
\includegraphics[width=1\columnwidth]{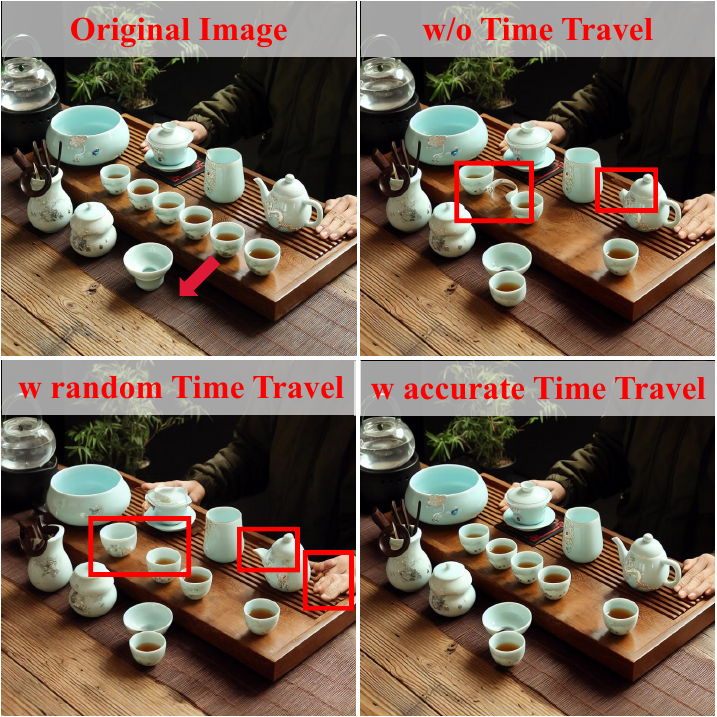}
\end{minipage}
\centering
\caption{
Visualization of editing without time travel, with random time travel, and with our accurate time travel.
}
\label{abs_travel} 
\end{figure}

\begin{figure*}[t]
\centering
\small 
\begin{minipage}[t]{\linewidth}
\centering
\includegraphics[width=.93\columnwidth]{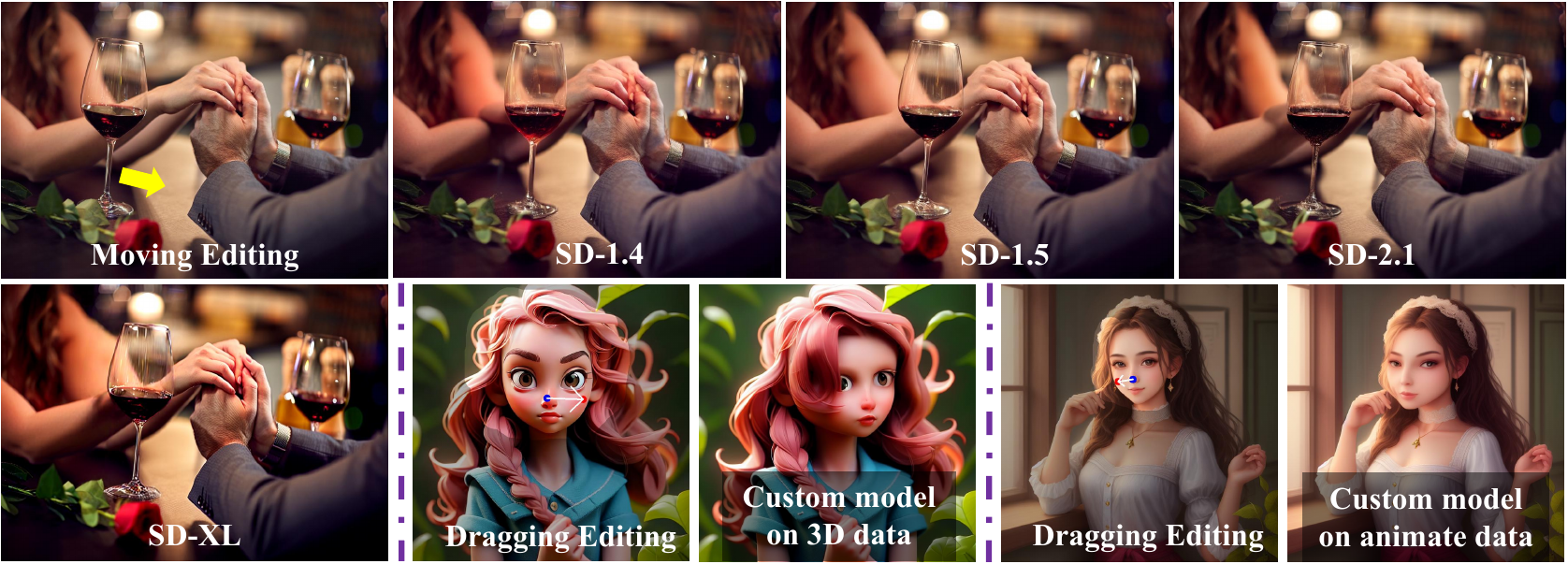}
\end{minipage}
\centering
\caption{
Visualization of editing results on different diffusion models.
}
\label{anything}
\end{figure*}

\subsection{Ablation Study}
\label{sec_abs}
In this part, we study the effectiveness of some components in our DiffEditor.

\noindent \textbf{Image prompt.} Image prompt provides a detailed description of the editing content in our editing pipeline. We conduct an experiment in Fig.~\ref{abs_ibm} to demonstrate its effectiveness. First, we apply it in the pure DDIM inversion and then reconstruction. The result in the first row presents that DDIM inversion based on the text prompt exhibits noticeable distortions and is unstable. After using our image prompt, DDIM inversion can reconstruct stable and high-fidelity results. In the second row of Fig.~\ref{abs_ibm}, we show the editing with and without the image prompt. It can be seen that the image prompt provides a better generation prior for editing content, reducing the probability of distortion. 

\noindent \textbf{Regional gradient guidance.} To rectify the interference between different gradient guidance, we use Eq.~\ref{l_gradient} to produce the guidance. We demonstrate its effectiveness in Fig.~\ref{abs_gradient} by only using $\mathbf{m}_{edit}\cdot \mathcal{E}_{edit}$ and $\mathcal{E}_{edit}$. Note that we remove the content consistency guidance $\mathcal{E}_{content}$ to highlight the interference. The results show that if the actuating range of $\mathcal{E}_{edit}$ is not constrained, the editing gradient will have an impact on the consistency of some unrelated areas, \textit{e.g.}, distortion of the fingers in the background. After applying regional constraints, the content of the background part has better consistency even without $\mathcal{E}_{content}$.

\noindent \textbf{Time travel.} Time travel is used to build recurrent guidance in a single diffusion time step, thereby refining the editing effect. We present its effectiveness in Fig.~\ref{abs_travel}. As can be seen, the editing result in the complex scenario has distortions without the time travel strategy. If using the random time travel (\textit{i.e.}, $\mathbf{z}_t\sim \mathcal{N}(\sqrt{1-\beta_{t-1}}\mathbf{z}_{t-1}, \beta_{t-1}\mathbf{I})$), the randomness will affect the consistency between the editing result and the original image. After adopting our accurate time travel, the editing quality is improved.

\subsection{Generalization of Different Components}
Except image prompt encoder that requires a specific SD model, other components of our method are designed based on diffusion theory, giving them good generalization. Fig.~\ref{anything} shows editing results of these components on different versions of SD and custom SD tuned on specific data. It shows the promising generalization of these components on different diffusion models. The reason we employ SD-1.5 is that the content of editing results mostly comes from existing images, thus the dependence on generation ability is relatively small. Large generative models, like SDXL, produce high computational costs in gradient guidance. Hence, we use the efficient SD-1.5 as the base model.

\section{Conclusion}
In this paper, we aim to rectify two issues in diffusion-based fine-grained image editing: (1) in complex scenarios, editing results often lack editing accuracy and exhibit unexpected artifacts; (2) lack of flexibility to harmonize editing operation, \textit{e.g.}, imagine new content. In our solution, we introduce the image prompt into fine-grained image editing, which can provide a more detailed content description for the edited image and reduce the probability of distortion. This method can be plugged into various fine-grained image editing tasks without task-specific training. To improve the editing flexibility, we propose a regional SDE strategy to inject randomness into the editing area while maintaining content consistency in other areas. Furthermore, we introduce regional score-based gradient guidance and a time travel strategy into the editing process to improve the editing quality further. Extensive experiments demonstrate that our method can achieve promising performance in various fine-grained image editing tasks, \textit{i.e.}, object moving, resizing, pasting, appearance replacing, and content dragging. The complexity is also reduced compared with existing diffusion-based methods. 

\noindent \textbf{Limitations} Although our method improves the flexibility of diffusion-based image editing and reduces the probability of distortion, editing difficulties still exist in some scenarios that require a large amount of content imagination, such as rotating a car by dragging its front. We think that this is due to the base model SD. It has a diverse generation space but lacks 3D perception of individual objects. In our future work, we will enhance the editing capabilities of diffusion models in this regard.
{
    \small
    \bibliographystyle{ieeenat_fullname}
    \bibliography{main}
}


\end{document}